\documentclass[10pt]{article} 
\usepackage[accepted]{rlc}
\usepackage{amssymb}            
\usepackage{mathtools}          
\usepackage{mathrsfs}           
\mathtoolsset{showonlyrefs}     
\usepackage{graphicx}           
\usepackage{subcaption}         
\usepackage[space]{grffile}     
\usepackage{url}                
\usepackage{natbib}
\usepackage{hyperref}
\usepackage{amsmath,amsfonts}
\usepackage{booktabs}
\usepackage{paralist}
\usepackage{todonotes}
\usepackage[ruled,lined,linesnumbered]{algorithm2e}
\DontPrintSemicolon
\newtheorem{definition}{Definition}

\SetKwComment{tcc}{$\triangleright$~}{}
\SetCommentSty{normalfont}
\SetKwInput{KwRequire}{Require}
\SetNlSty{}{}{:}
\title{MOSEAC: Streamlined Variable Time Step Reinforcement Learning}

\author{Dong Wang \\
    dong-1.wang@polymtl.ca \\
    Department of Computer Engineering\\
    Polytechnique Montr\'eal
    \And
    Giovanni Beltrame \\
    giovanni.beltrame@polymtl.ca\\
    Department of Computer Engineering \\
    Polytechnique Montr\'eal}

\begin{document}

\maketitle

\begin{abstract}
  Traditional reinforcement learning (RL) methods typically employ a fixed
  control loop, where each cycle corresponds to an action. This rigidity poses
  challenges in practical applications, as the optimal control frequency is
  task-dependent. A suboptimal choice can lead to high computational demands and
  reduced exploration efficiency. Variable Time Step Reinforcement Learning
  (VTS-RL) addresses these issues by using adaptive frequencies for the control
  loop, executing actions only when necessary. This approach, rooted in reactive
  programming principles, reduces computational load and extends the action
  space by including action durations. However, VTS-RL's implementation is often
  complicated by the need to tune multiple hyperparameters that govern
  exploration in the multi-objective action-duration space (i.e., balancing task
  performance and number of time steps to achieve a goal). To overcome these
  challenges, we introduce the Multi-Objective Soft Elastic Actor-Critic
  (MOSEAC) method. This method features an adaptive reward scheme that adjusts
  hyperparameters based on observed trends in task rewards during training. This
  scheme reduces the complexity of hyperparameter tuning, requiring a
  single hyperparameter to guide exploration, thereby simplifying the learning
  process and lowering deployment costs. We validate the MOSEAC method through
  simulations in a Newtonian kinematics environment, demonstrating high task and
  training performance with fewer time steps, ultimately lowering energy
  consumption. This validation shows that MOSEAC streamlines RL algorithm
  deployment by automatically tuning the agent control loop frequency using a
  single parameter. Its principles can be applied to enhance any RL algorithm,
  making it a versatile solution for various applications.
\end{abstract}

\section{Introduction}
\label{sec:introdction}

Model-free deep reinforcement learning (RL) algorithms have achieved significant
success in various domains, including gaming \citep{liu2021deep, ibarz2021train}
and robotic control \citep{akalin2021reinforcement, singh2022reinforcement}.
Traditional RL methods typically rely on a fixed control loop where the agent
takes actions at predetermined intervals, e.g. every 0.1 seconds (10 Hz).
However, this approach introduces significant limitations: fixed-rate control
can lead to stability issues and high computational demands, especially in
dynamic environments where the optimal action frequency may vary over time.

To address these limitations, Variable Time Step Reinforcement Learning (VTS-RL)
was recently introduced, based on reactive programming principles
\citep{majumdar2021paracosm, bregu2016reactive}. The key idea behind VTS-RL is
to \emph{execute control actions only when necessary}, thereby reducing
computational load, as well as expanding the action space including variable
action durations.
For instance, in an autonomous driving scenario,
VTS-RL enables the agent to adapt its control strategy based on the situation,
using a low frequency in stable conditions and a high frequency in
more complex situations \citep{wang2024deployable}.

Two notable VTS-RL algorithms are Soft Elastic Actor-Critic (SEAC)
\citep{wang2024deployable} and Continuous-Time Continuous-Options (CTCO)
\citep{karimi2023dynamic}, which simultaneously learn actions and their
durations.

CTCO employs continuous-time decision-making and flexible option durations,
improving exploration and robustness in continuous control tasks. However, its
effectiveness can be hampered by the need to tune several hyperparameters, and
the fact that tasks can take a long time to complete as CTCO does not optimize
for task duration.

In contrast, SEAC adds reward terms for task energy (i.e. the numbers of
actions performed) and task time (i.e. the time needed to complete a task),
showing effectiveness in time-restricted environments like a racing video game
\citep{wang2024reinforcement}. However, SEAC requires careful tuning of its
hyperparameters (balancing task, energy, and time costs) to avoid performance
degradation.

SEAC and CTCO's brittleness to hyperparameter settings poses a challenge for
users aiming to fully leverage their potential. To mitigate this issue, we
propose the Multi-Objective Soft Elastic Actor-Critic (MOSEAC) algorithm based
on SEAC \citep{wang2024deployable}. MOSEAC adds action
durations to the action space and adjusts hyperparameters based on observed
trends in task rewards during training, leaving a single hyperparameter to be
set to guide the exploration. We evaluate MOSEAC in a Newtonian kinematics
simulation environment, comparing it CTCO, SEAC, and fixed-frequency SAC,
showing that MOSEAC is stable, accelerates training, and has high final task
performance. Additionally, our hyperparameter setting approach can be broadly
applied to any continuous action reinforcement learning algorithm, such as Twin
Delayed Deep Deterministic (TD3) \citep{fujimoto2018addressing} or Proximal
Policy Optimization (PPO) \citep{schulman2017proximal}. This adaptability
facilitates the transition from fixed-time step to variable-time step
reinforcement learning, significantly expanding the scope of its practical
application.

\section{Related Work}
\label{sec:related_work}


The importance of action duration in reinforcement learning algorithms has been
severely underestimated for a long time. In fact, if reinforcement learning
algorithms are to be applied to the real world, it is an essential factor,
impacting an agent's exploration capabilities \citep{amin2020locally,
  park2021time}. High frequencies may reduce exploration efficiency in some
cases, but are needed in others such as environments with delays
\citep{bouteiller2021reinforcement}. \citet{wang2024reinforcement,
  karimi2023dynamic} show how various frequencies impact learning, with high
frequencies potentially impeding convergence. Thus, a dynamically optimal
control frequency, which adjusts in real time based on reactive principles,
could enhance performance and adaptability.

\citet{sharma2017learning} introduced a concept similar to variable control
frequencies, developing a framework for repetitive action reinforcement
learning. This model allows an agent to perform identical actions over
successive states, combining them into a larger action. This method has
intrigued researchers, especially in gaming \citep{metelli2020control,
  lee2020reinforcement}, but fails to account for physical properties or reduce
computational demands, making its practical application challenging.

Additionally, \citet{chen2021varlenmarl} attempted to modify the traditional
``control rate'' by integrating actions such as ``sleep'' to reduce activity
periods ostensibly. However, this approach still mandates fixed-frequency checks
of system status, which does not effectively diminish the computational load as
anticipated. These instances underscore the ongoing challenges and the nascent
stage of effectively integrating variable control frequencies and repetitive
behaviors into real-world applications, highlighting a critical gap between
theoretical innovation and practical efficacy.

\section{Algorithm Framework}
\label{sec:algorithm}

To reduce hyperparameter dimensions, we combine SEAC's
\citep{wang2024deployable} hyperparameters balancing task, energy, and
time rewards using a simple multiplication method. We then apply an adaptive
adjustment method on the remaining hyperparameters.
Like SEAC, our reward equation includes components for:
\begin{compactitem}
\item Quality of task execution (the standard RL reward),
\item Time required to complete a task (important for varying action durations),
  and
\item Energy (the number of time steps, a.k.a. the number of actions taken, which
  we aim to minimize).
\end{compactitem}

\begin{definition}\label{def:reward_policy}
	\begin{equation}
	R = \alpha_{m} R_t R_{\tau} - \alpha_{\varepsilon}
	\end{equation}
	where $R_{t}$ is the task reward; $R_{\tau}$ is a time dependent term.
  
  $\alpha_{m}$ is a weighting factor used to modulate the magnitude of the
  reward. Its primary function is to prevent the reward from being too small,
  which could lead to task failure, or too large, which could cause reward
  explosion, ensuring stable learning.
  
  $\alpha_{\varepsilon}$ is a penalty parameter applied at each time step to
  impose a cost on the agent’s actions. This parameter gives a fixed cost to the
  execution of an action, thereby discouraging unnecessary ones. In practice
  $\alpha_{\varepsilon}$ promotes the completion of a task using fewer time
  steps (remember that time steps have variable duration), i.e. it reduces the
  energy used by the control loop of the agent.
  
	We determine the optimal policy $\pi^*$, which maximizes the reward $R$.
\end{definition}


The reward as designed, minimizes both energy cost (number of steps) and the
total time to complete the task through $R_{\tau}$. By scaling the
task-specific reward based on action duration with $\alpha_m$, agents are
motivated to complete tasks using fewer actions:

\begin{definition}\label{def:time_reward}
	The remap relationship between action duration and reward
	\begin{equation}
	R_{\tau} = t_{min} / t, \quad R_{\tau} \in [t_{min}/t_{max}, 1] 
	\end{equation}
	where $t$ is the duration of the current action, $t_{min}$ is the minimum duration
  of an action (strictly greater than 0), and $t_{max}$ is the maximum duration of
  an action.
\end{definition}


To automatically set $\alpha_{m}$ and $\alpha_{\varepsilon}$ to optimal values,
we adjust them dynamically during training. Based on
\citet{wang2024reinforcement}'s experience, we increase $\alpha_{m}$ and
decrease $\alpha_{\varepsilon}$ to mitigate convergence issues, specifically the
problem of sparse rewards caused by a suboptimal set of a large
$\alpha_{\varepsilon}$ and a small $\alpha_{m}$. This adjustment ensures that
rewards are appropriately balanced, facilitating learning and convergence.

These parameters are adjusted based on observed trends in task rewards: if the
average reward is declining (see Definition \autoref{def:linregress}), we
increase $\alpha_{m}$ and decrease $\alpha_{\varepsilon}$, linking them with a
sigmoid function, ensuring a balanced reward structure that promotes convergence
and efficient learning. To guarantee stability, we ensure the change is
monotonic, forcing a uniform sweep of the parameter space.


\begin{definition}\label{def:r_e}
	The relationship between $\alpha_{\varepsilon}$ and $\alpha_{m}$ is
	\begin{equation}
	\text{{$\alpha_{\varepsilon}$}} = 0.2 \cdot \left(1 - \frac{1}{1 + e^{-\alpha_m + 1}}
	\right)
	\end{equation}

  Based on \citet{wang2024deployable}'s experience, we establish a mapping
  relationship between the two parameters: when the initial value of
  $\alpha_{m}$ is 1.0, the initial value of $\alpha_{\varepsilon}$ is 0.1. As
  $\alpha_{m}$ increases, $\alpha_{\varepsilon}$ decreases, but never falls
  below 0.
\end{definition}

To determine the trend on the average reward, we perform a linear regression
across the current training episode and compute the slope of the resulting line:
if it is negative, the reward is declining.

\begin{definition}\label{def:linregress}
	The slope of the average reward ($k_R$) is:
	\begin{equation}
	k_R(R_a) = \frac{n \sum_{i=1}^{n} (i \cdot {R_a}_i) - \left( \sum_{i=1}^{n} i \right) \left( \sum_{i=1}^{n} {R_a}_i \right)}{n \sum_{i=1}^{n} i^2 - \left( \sum_{i=1}^{n} i \right)^2}
	\end{equation}
	where $n$ is the total number of data points collected across the update 
	interval ($k_{update}$ in Algorithm \autoref{algo:SEAC}). The update 
	interval is a hyperparameter that determines the frequency of updates for 
	these neural networks used in the actor and critic policies, occurring 
	after every $n$ episode \citep{sutton2018reinforcement}. $R_a$ represents 
	the list of average rewards $({R_a}_1,{R_a}_2,...,{R_a}_n)$ during training.
	 Here, an average reward ${R_{a}}_i$ is calculated across one episode.
\end{definition}

After observing a downward trend in the average reward during the training process, we introduced the hyperparameter $\psi$ to dynamically adjust $\alpha_m$ as defined in Definition \autoref{def:monotonic}.

\begin{definition}\label{def:monotonic}
	We adaptively adjust the reward every $k_{update}$ episodes. When $k_R < 0$:
	\begin{equation}
		\alpha_{m} = \alpha_{m} + \psi
	\end{equation}
	where, $\psi$ is the only additional hyperparameter required by our MOSEAC to tune the reward equation during training. Additionally, $\alpha_{\varepsilon}$ is adjusted as described in Definition \autoref{def:r_e}.
\end{definition}

Algorithm \autoref{algo:SEAC} shows the pseudocode of MOSEAC. Convergence
analysis is provided in \autoref{sec:appendix_conv}.

\begin{algorithm}[H]\label{algo:SEAC}
	\SetAlgoLined
	\KwRequire{a policy $\pi$ with a set of parameters $\theta$, $\theta^{'}$, critic parameters $\phi$, $\phi^{'}$, variable time step environment model $\Omega$, learning-rate $\lambda_p$, $\lambda_q$, reward buffer $\beta_r$, replay buffer $\beta$.}
			
	Initialization $i = 0$, $t_i = 0$, $\beta_r = 0$, observe $S_0$\\
	\While{$t_i \leq t_{max}$}{
		\For{$i \leq k_{length} \vee Not \, Done$}{
			$A_i, D_i = \pi_{\theta}(S_{i})$ \hfill {$\rightarrow$ simple action and its duration}\\
			$S_{i+1}, R_i = \Omega(A_i, D_i)$ \hfill {$\rightarrow$ compute reward with Definition \ref{def:reward_policy} and \ref{def:time_reward}}\\
			
			$i \leftarrow i+1$
		}
		$\beta_{r} \leftarrow 1/i \times \sum_{0}^{i} R_i$ \hfill {$\rightarrow$ collect the average reward for one episode}\\
		$\beta \leftarrow S_{0 \sim i}, \, A_{0 \sim i}, \, D_{0 \sim i}, \, R_{0 \sim i}, \, S_{1 \sim i+1}$ \\
		$i = 0$\\
		$t_{i} \leftarrow t_{i} + 1$ \\ 
		\If{$t_{i} \geq k_{init} \quad \& \quad t_{i} \mid k_{update}$}{
			$Sample \, S, \, A, \, D, \, R, \, S^{'} from (\beta)$ \\
			$\phi \leftarrow \phi - \lambda_q\nabla_{\delta}\mathcal{L}_{Q}(\phi, \, S, \, A, \, D, \, R, \, S^{'})$ \hfill {$\rightarrow$ critic update}\\
			$\theta \leftarrow \theta - \lambda_p\nabla_{\theta}\mathcal{L}_{\pi}(\theta, \, S, \, A, \, D, \, \phi)$ \hfill {$\rightarrow$ actor update}\\
			\If{$k_R(\beta_r)$}{
				$\alpha_{m} \leftarrow \alpha_{m} + \psi$ \hfill {$\rightarrow$ see Definition \autoref{def:linregress} for $k_R$}\\
				$\alpha_{\varepsilon} \leftarrow F_{update}(\alpha_{m})$ \hfill {$\rightarrow$ update $\alpha_{m}, \alpha_{\varepsilon}$ followed Definition \ref{def:r_e}}\\}
			$\beta_{r} = 0$ \hfill {$\rightarrow$ Re-record average reward values under new hyperparameters}\\
		}
		Perform soft-update of $\phi^{'}$ and $\theta^{'}$
	}
	\caption{Multi-Objective Soft Elastic Actor and Critic}
\end{algorithm}

Here, $t_{max}$ represents maximum training steps 
\citep{sutton2018reinforcement}; $k_{length}$ is maximum exploration steps per 
episode \citep{sutton2018reinforcement}; $k_{init}$ is steps in the initial 
random exploration phase \citep{sutton2018reinforcement}. The reward $R_i$ is 
calculated as $R(S_i, A_i, D_i)$, where $D_i$ lies within $[T_{min}, T_{max}]$, 
representing action duration.

Our algorithm optimizes multiple objectives using a streamlined, weighted
strategy. Unlike Hierarchical Reinforcement Learning (HRL)
\citep{dietterich2000hierarchical, li2019hierarchical}, which seeks Pareto
optimality \citep{kacem2002pareto, monfared2021pareto} with layered reward
policies, our method simplifies the approach. We emphasize user-friendliness and
computational efficiency, making our strategy easily adaptable to
various algorithms.



Apart from a series of hyperparameters inherent to RL that need adjustment, such
as learning rate, $t_{max}$, $k_{update}$, etc., $\psi$ is the only
hyperparameter that requires tuning in MOSEAC.

However, one limitation of our approach is that high $\psi$ values can lead to 
issues such as reward explosion. While a small $\psi$ value generally avoids
significant problems and guides $\alpha_{m}$ to its optimal value, it requires
extended training periods. Determining the appropriate $\psi$ value remains a
critical optimization point in our method. We recommend using the pre-set $\psi$
value provided in our implementation, thus reducing the need for additional
parameter adjustments. If users encounter problems such as reward explosion, it
is advisable to reduce the $\psi$ value appropriately. Our ongoing research aims
to mitigate further the risks associated with $\psi$, enhancing the algorithm’s
reliability and robustness.


\section{Empirical Analysis}
\label{sec:results}
We conduct six experiments involving MOSEAC, SEAC, CTCO, and SAC algorithms (at
20 Hz and 60 Hz) within a Newtonian kinematics environment. A detailed
description of the environment appears in \autoref{sec:appendix_env}. These
tests are performed on a computer equipped with an Intel Core i5-13600K and a
Nvidia RTX 4070 GPU. The operating system is Ubuntu 22.04 LTS. For the
hyperparameter settings of MOSEAC, please refer to
\autoref{sec:appendix_parameters}\footnote{Our 
	\href{https://github.com/alpaficia/MOSEAC}{code} is public available.}

\begin{figure}[htbp]
	\centering
	\begin{subfigure}{.49\textwidth}
		\centering
		\includegraphics[width=\linewidth]{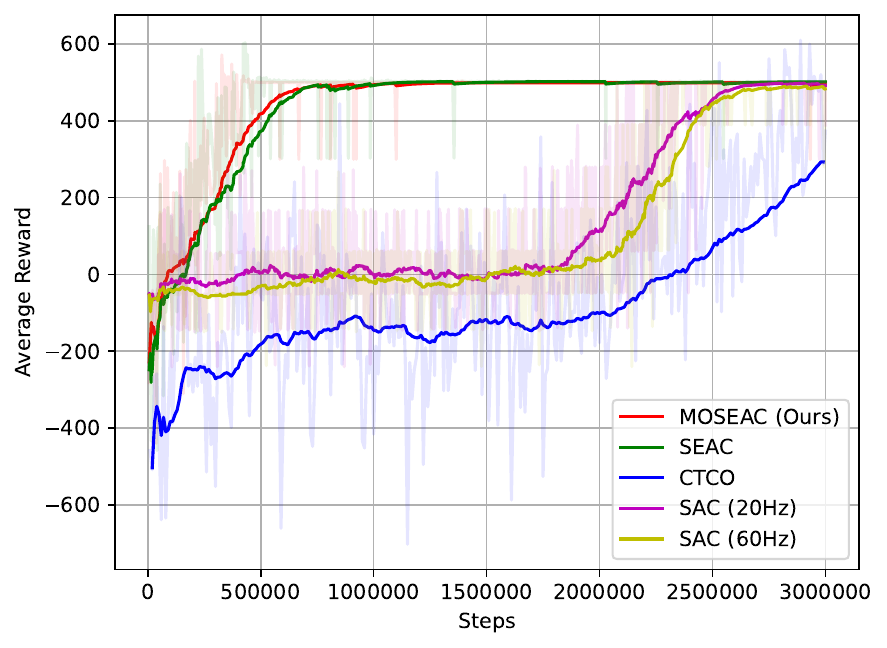}
		\caption{Average returns of 5 reinforcement learning algorithms in 
			3M steps during the training.}
		\label{fig:average_reward}
	\end{subfigure}\hfill
	\begin{subfigure}{.49\textwidth}
		\centering
		\includegraphics[width=\linewidth]{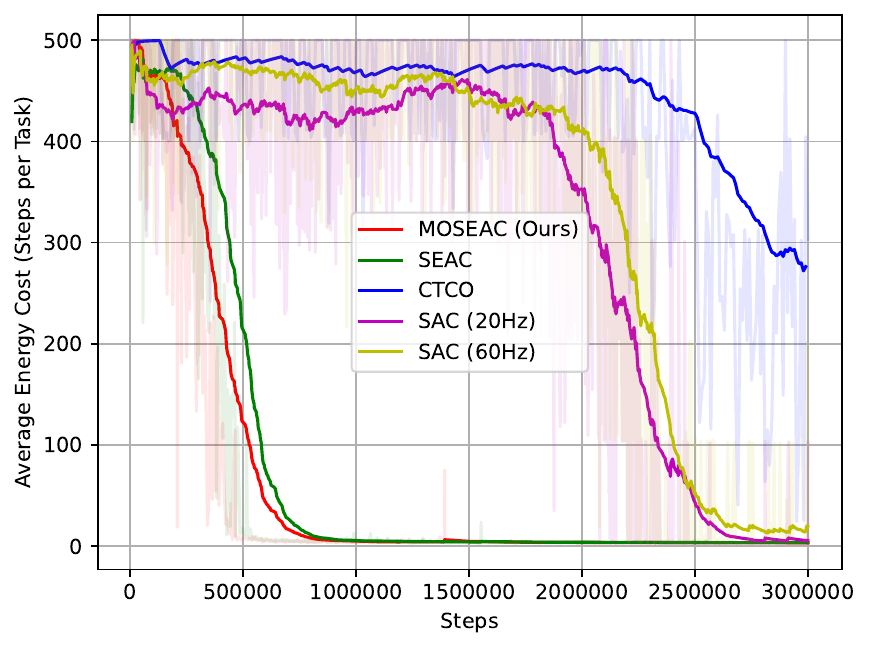}
		\caption{Average Energy cost of 5 reinforcement learning algorithms in 
			3M steps during the training.}
		\label{fig:average_energy_cost}
	\end{subfigure}
	\caption{Result graphs of five reinforcement learning algorithms.}
	\label{fig:training}
\end{figure}

\autoref{fig:training} illustrates the result of training process. In
\autoref{fig:average_reward}, the average returns align with our expectations,
showing that CTCO is sensitive to the choice of action duration range, which
affects its discount factor $\gamma$. During the training process $\gamma$
tended to be very small, compromising CTCO's long-term planning. Overall, CTCO
performs worse than the two fixed-control frequency SAC baselines in this
environment. 
In contrast, MOSEAC trains slightly faster than SEAC, as shown in
\autoref{fig:average_energy_cost}. 


\begin{figure}[htbp]
	\centering
	\begin{subfigure}{.49\textwidth}
		\centering
		\includegraphics[width=\linewidth]{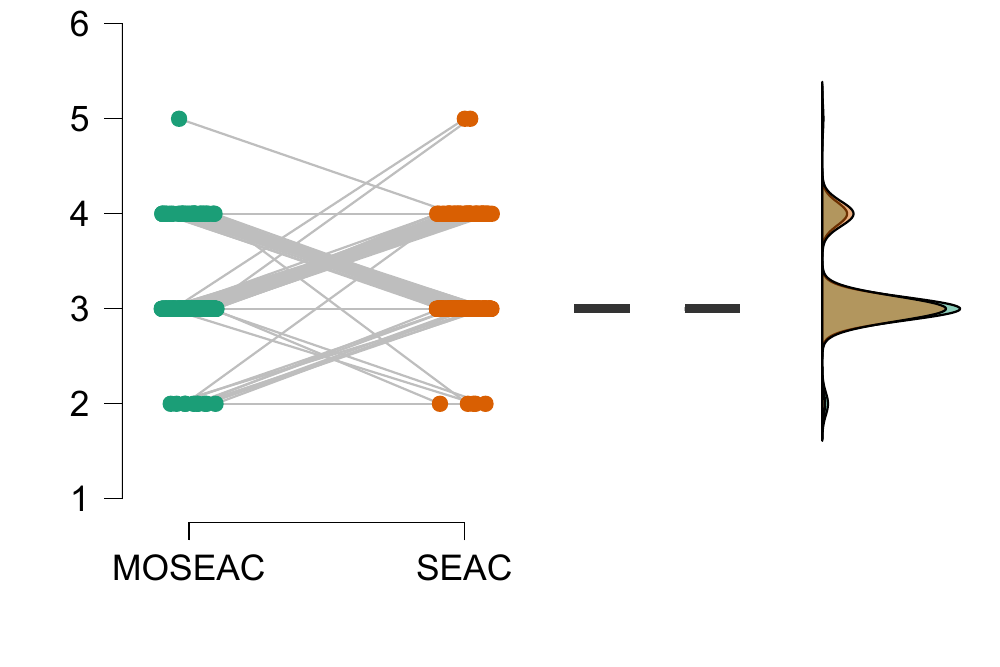}
		\caption{The energy cost (counted by numbers of steps) figure 
			for 300 different tasks. }
		\label{fig:energy}
	\end{subfigure}\hfill
	\begin{subfigure}{.49\textwidth}
		\centering
		\includegraphics[width=\linewidth]{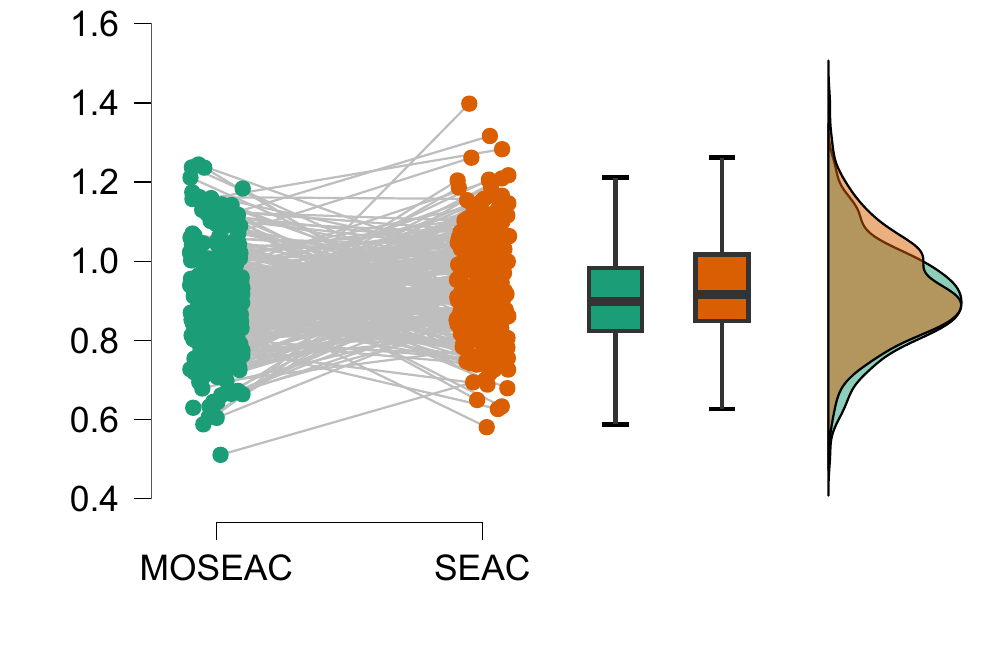}
		\caption{The time cost figure for 300 different tasks.}
		\label{fig:time}
	\end{subfigure}
	\caption{Comparison figure of energy consumption and time performance of 
		300 different tasks. We use the same random seed to ensure that both 
		have the same tasks.}
	\label{fig:raingraph}
\end{figure}

We compared MOSEAC's task performance with the best-performing SEAC after
training. Due to the poor performance of CTCO and SAC, we excluded them from the
analysis. \autoref{fig:raingraph} presents performance metrics related to energy
consumption and task duration. Specifically, \autoref{fig:energy} shows the
average energy consumption (measured in steps) for 300 tasks, and
\autoref{fig:time} compares average task durations. Data show that MOSEAC's
average energy and time costs are lower than SEAC's.

Wilcoxon signed-rank test (chosen due to the lack of normality in the data
distribution) indicates that MOSEAC's energy cost is lower than SEAC's (z =
-1.823, p = 0.020), where MOSEAC has a mean energy cost of 3.120 (SD = 0.424),
compared to SEAC's mean of 3.193 (SD = 0.451). Detailed statistical results are
provided in \autoref{sec:appendix_statistic}. Similarly, MOSEAC's time cost is
significantly lower than SEAC's (z = -2.669, p = 0.004), with MOSEAC having a
mean time cost of 0.905 (SD = 0.125), compared to SEAC's mean of 0.935 (SD =
0.127).

The improved performance of MOSEAC over SEAC can be attributed to its reward
function. While SEAC’s reward function is linear, combining task reward, energy
penalty, and time penalty independently, MOSEAC introduces a multiplicative
relationship between task reward and time-related reward. This non-linear
interaction enhances the reward signal, particularly when both task performance
and time efficiency are high, and naturally balances these factors. By keeping
the energy penalty separate, MOSEAC maintains flexibility in tuning without
complicating the relationship between time and task rewards. This design allows
MOSEAC to more effectively guide the agent’s decisions, resulting in better
energy efficiency and task completion speed in practical applications.

\section{Conclusion and Future Work}
\label{sec:conclusion}

In this paper, we present MOSEAC, a VTS-RL algorithm with adaptive
hyperparameters that respond to observed reward trends during training. This approach
significantly reduces hyperparameter sensitivity and increases robustness,
improving data efficiency. Unlike other VTS-RL algorithms, our method does not
requires a single hyperparameter, thereby reducing learning and tuning costs when
transitioning from fixed to variable time steps. This establishes a strong
foundation for real-world applications. 

Our next objective is to implement our method in real-world robotics
applications, such as smart cars and robotic arms.

\clearpage
\bibliography{rlc24}
\bibliographystyle{rlc}

\clearpage
\appendix

\section{Convergence Analysis of MOSEAC}
\label{sec:appendix_conv}

This appendix aims to analyze the convergence of the Multi-Objective Soft 
Elastic Actor-Critic (MOSEAC) algorithm. Its action space includes a time 
dimension $D$ and the reward function is $R = \alpha_{m} R_{t} R_{\tau} - 
\alpha_{\varepsilon}$, with $\alpha_{m}$ and $\alpha_{\varepsilon}$ dynamically 
adjusted during training, details can be found in \autoref{sec:algorithm}.

\subsection{SAC Algorithm Overview}
In the standard SAC algorithm \citep{haarnoja2018soft}, the policy 
$\pi_\theta(a|s)$ selects action $a$, and updates the policy parameters 
$\theta$ and value function parameters $\phi$. The objective function is:
\begin{equation}
J(\pi_\theta) = \mathbb{E}_{(s, a) \sim \pi_\theta} \left[ Q^\pi(s, a) + 
\alpha \mathcal{H}(\pi_\theta(\cdot|s)) \right]
\end{equation}
where $J(\pi_\theta)$ is the objective function, $Q^\pi(s, a)$ is the 
state-action value function, $\alpha$ is a temperature parameter controlling 
the entropy term, and $\mathcal{H}(\pi_\theta(\cdot|s))$ is the entropy of the 
policy.

\subsection{Incorporating Time Dimension and Our Reward Function}
When the action space is extended to include $D$ and the reward function is 
modified to $R = \alpha_{m} R_{t} R_{\tau} - \alpha_{\varepsilon}$, where 
$\alpha_{m} \geq 0$, $0 < R_{\tau} \leq 1$, and $\alpha_{\varepsilon}$ is a small 
positive constant, the new objective function becomes:
\begin{equation}
J(\pi_\theta) = \mathbb{E}_{(s, a, D) \sim \pi_\theta} \left[ Q^\pi(s, a, D) + 
\alpha \mathcal{H}(\pi_\theta(\cdot|s)) \right]
\end{equation}
where $Q^\pi(s, a, D)$ is the extended state-action value function with time 
dimension $D$, and $R_{t}$ and $R_{\tau}$ are components of the reward 
function, see Definition \autoref{def:reward_policy}.

\subsection{Policy Gradient}
With our reward function, the policy gradient is:
\begin{equation}
\nabla_\theta J(\pi_\theta) = \mathbb{E}_{\pi_\theta} \left[ \nabla_\theta \log 
\pi_\theta(a, D|s) \left( Q^\pi(s, a, D) \cdot (\alpha_{m} \cdot R_{\tau}) - 
\alpha_{\varepsilon} \right) \right]
\end{equation}
where $\nabla_\theta J(\pi_\theta)$ is the gradient of the objective function 
with respect to the policy parameters $\theta$.

\subsection{Value Function Update}
The value function update, incorporating the time dimension $D$ and our reward 
function, is:
\begin{equation}
L(\phi) = \mathbb{E}_{(s, a, D, r, s')} \left[ \left( Q_\phi(s, a, D) - \left( 
r + \gamma \mathbb{E}_{(a', D') \sim \pi_\theta} [V_{\bar{\phi}}(s') - \alpha 
\log \pi_\theta(a', D'|s')] \right) \right)^2 \right]
\end{equation}
where $L(\phi)$ is the loss function for the value function update, $r$ is the 
reward, $\gamma$ is the discount factor, and $V_{\bar{\phi}}(s')$ is the target 
value function.

\subsection{Policy Parameter Update}
The new policy parameter $\theta$ update rule is:
\begin{equation}
\theta_{k+1} = \theta_k + \beta_k \mathbb{E}_{s \sim D, (a, D) \sim \pi_\theta} 
\left[ \nabla_\theta \log \pi_\theta(a, D|s) \left( Q_\phi(s, a, D) \cdot 
(\alpha_{m} \cdot R_{\tau}) - \alpha_{\varepsilon} - V_{\bar{\phi}}(s) + \alpha 
\log \pi_\theta(a, D|s) \right) \right]
\end{equation}
where $\beta_k$ is the learning rate at step $k$.

\subsection{Dynamic Adjustment and Convergence Analysis}
To analyze the impact of dynamically adjusting $\alpha_{m}$ and 
$\alpha_{\varepsilon}$, we assume:

\begin{enumerate}
	\item {\bf{Dynamic Adjustment Rules:}}\\
	- $\alpha_{m}$ increases monotonically by a small increment $\psi$ if the 
	reward trend decreases over consecutive episodes (See Definition 
	\autoref{def:linregress}).\\
	- $\alpha_{\varepsilon}$ decreases with the Definition \autoref{def:r_e}.
	
	\item{\bf{Learning Rate Conditions:}}\\
	The learning rates $\alpha_k$ and $\beta_k$ must satisfy the following 
	conditions \citep{konda1999actor}:
	\begin{equation}
	\sum_{k=0}^{\infty} \alpha_k = \infty, \quad \sum_{k=0}^{\infty} 
	\alpha_k^2 < \infty
	\end{equation}
	\begin{equation}
	\sum_{k=0}^{\infty} \beta_k = \infty, \quad \sum_{k=0}^{\infty} 
	\beta_k^2 < \infty
	\end{equation}
\end{enumerate}

\subsection{Unbiased Estimation}
Assuming the critic estimates are unbiased:
\begin{equation}
\mathbb{E}[Q_\phi(s, a, D) \cdot (\alpha_{m} \cdot R_{\tau}) - 
\alpha_{\varepsilon}] = Q^\pi(s, a, D) \cdot (\alpha_{m} \cdot R_{\tau}) - 
\alpha_{\varepsilon}
\end{equation}
Since $R_{\tau}$ is a positive number within [0, 1], its effect on 
$Q^\pi(s, a, D)$ is linear and does not affect the consistency of the policy 
gradient.

\subsection{Impact Evaluation}
\begin{enumerate}
	\item{\bf{Positive Scaling:}}\\
	As $R_{\tau}$ is always a positive number less than or equal to 1, and 
	$\alpha_{m} \geq 0$, it only scales the reward, not altering its sign. This 
	scaling does not change the direction of the policy gradient but affects 
	its magnitude.
	
	\item{\bf{Small Offset:}}\\
	$\alpha_{\varepsilon}$ is a small constant used to accelerate training. This 
	small offset does not affect the direction of the policy gradient but 
	introduces a minor shift in the value function, which does not alter the 
	overall policy update direction.
\end{enumerate}

\subsection{Conclusion}
Under the aforementioned conditions, the modified SAC algorithm with the time 
dimension $D$ and the new reward function $R = \alpha_{m} R_{t} R_{\tau} - 
\alpha_{\varepsilon}$, which we called MOSEAC, will converge to a local optimum, 
i.e. \citep{sutton2018reinforcement},
\begin{equation}
\lim_{k \to \infty} \nabla_\theta J(\pi_\theta) = 0
\end{equation}
This implies that over time, the policy parameters will stabilize at a local 
optimum, maximizing the policy performance.
\clearpage

\section{Details of Newtonian kinematics environment}
\label{sec:appendix_env}

To validate our algorithm, we set up an autonomous driving environment based on 
Newtonian dynamics, including a random birth point, a random endpoint, and a 
random obstacle. Check Definition \autoref{def:state} for its state value 
information, Definition \autoref{def:action} for its action value information, 
and Definition \autoref{def:newton_formulas} for the physical formula it follows.

As Definition \ref{def:reward_policy}, the precise reward configuration for our 
environment are outlined in Table \ref{table:reward}.

\begin{definition}\label{def:newton_formulas}
	The Newton Kinematics formulas are:
	\begin{equation}
	D_{aim} = 1/2 \cdot (V_{aim}+V_{current}) \cdot T;
	\end{equation}
	\begin{equation}
	V_{aim} = V_{current} + AT;
	\end{equation}
	\begin{equation}
	F_{aim} = mA;
	\end{equation}
	\begin{equation}
	F_{true} = F_{aim} - f_{friction};
	\end{equation}
	\begin{align}
	f_{\text{friction}} &= \mu mg, \text{ if } F_{\text{aim}} > f_{\text{friction}} \land V_{\text{agent}} \neq 0 \\ 
	f_{\text{friction}} &= F_{\text{aim}}, \text{ if } F_{\text{aim}} \leq f_{\text{friction}} \land V_{\text{agent}} = 0 \notag
	\end{align}
	where $D_{aim}$ is the distance generated by the policy that the agent needs
	to move. $V_{agent}$ is the speed of the agent. $T$ is the time to complete
	the movement generated by the policy, $F_{aim}$ is the traction force necessary 
	for the agent to change to the aimed speed. $m$ is the mass of the agent, 
	$\mu$ is the friction coefficient, and $g$ is the acceleration of gravity.
\end{definition}

\begin{table}[htbp]
	\centering
	\caption{Reward Settings for The Newtonian Kinematics Environment}
	\begin{tabular}{llcl}
		\toprule
		\multicolumn{4}{c}{Reward Settings} \\
		\midrule
		Name                  & Value                  &                 & Annotation                 \\
		\midrule
		& $500.0$                   &                 & Reach the goal             \\
		r                     & $-500.0$                  &                 & Crash on an obstacle       \\
		& $D_{origin}-1.0 \times D_{goal}$   &                & $D_{origin}$: distance from start to goal; $D_{goal}$: distance to goal              \\
		$\alpha_m$            & $1.0$      &    &                     \\
		\bottomrule
	\end{tabular}
	\label{table:reward}
\end{table}

The state dimensions are shown in Definition \autoref{def:state}:

\begin{definition}\label{def:state}
	
	The positions of the endpoint and obstacle are randomized in each episode.
	
	\begin{equation}
	S_t=(Pos, Obs, Goal, Speed, Time, Force)
	\end{equation}
	
	Where:\\
	$Pos$ = Position Data of The Agent in X and Y Direction, \\
	$Obs$ = Position Data of The Obstacle in X and Y Direction, \\
	$Goal$ = Position Data of The Goal in X and Y Direction,\\
	$Speed$ = Spped Data of The Agent for Current Step,\\
	$Time$ = The Duration data of The Agent to Execute The Last Time Step,\\
	$Force$ = Force Data of The Agent for Last Time Step in X and Y Direction.
\end{definition}

The Spatial Information of our environment are shown in \autoref{table:state}:
\begin{table}[htbp]
	\centering
	\caption{Details of The Simple Newtonian Kinematics Gymnasium Environment}
	\begin{tabular}{llcl}
		\toprule
		\multicolumn{4}{c}{Environment details} \\
		\midrule
		Name                  & Value                  &                 & Annotation                 \\
		\midrule
		Action dimension     & $3$                   &                 & \\
		Range of speed       & $[-2, 2]$                  &                 &m/s   \\
		Action Space         & $[-100.0, 100.0]$   &                &Newton  \\
		Range of time        & $[0.01, 1.0]$      &     &second  \\
		State dimension      & $11$      &    & Task gain factor                    \\
		World size           & $(2.0, 2.0)$      &    & in meters                    \\
		Obstacle shape       & Round      &    & Radius: 5cm                   \\
		Agent weight         & 20      &    & in $Kg$                   \\
		Gravity factor       & 9.80665      &    & in $m/s^2$                   \\
		Static friction coeddicient      & 0.6      &    &                   \\
		\bottomrule
	\end{tabular}
	\label{table:state}
\end{table}

The action dimensions are shown in Definition \autoref{def:action}:

\begin{definition}\label{def:action}
	Action
	
	Each action should have a corresponding execution duration.\\
	\begin{equation}
	a_t=(D_{t}, A_{fx}, A_{fy})
	\end{equation}
	
	Where:\\
	$D_{t}$ = The duration of the Agent to implement current action, \\
	$A_{fx}$ = The Force of the Agent in X Coordinate, \\
	$A_{fy}$ = The Force of the Agent in Y Coordinate.
\end{definition}

\section{Hyperparameters of MOSEAC}
\label{sec:appendix_parameters}

\begin{table}[htbp]
	\centering
	\begin{tabular}{llcl}
		\toprule
		\multicolumn{4}{c}{Hyperparameter sheet of MOSEAC} \\
		\midrule
		Name                  & Value                  &                 & Annotation                 \\
		\midrule
		Total steps       & $3e6$                   &                 &  \\
		$\gamma$          & $0.99$                  &          & Discount factor  \\
		Net shape         & $(256, 256)$   &                &  \\
		batch\_size       & $256$      &     &  \\
		a\_lr             & $3e-5$      &    & Learning rate of Actor Network                    \\
		c\_lr             & $3e-5$      &    & Learning rate of Critic Network      \\
		max\_steps        & $500$      &    & Maximum steps for one episode                   \\
		$\alpha$          & $0.12$      &    &                    \\
		$\eta$            & $-3$      &    & Refer to SAC \citep{haarnoja2018soft2}                   \\
		min\_time         & $0.01$      &    & Minimum control duration, in seconds                   \\
		max\_time         & $1.0$      &    & Maximum control duration, in seconds                \\
		$\alpha_{m}$      & $1.0$      &    & Init value of $\alpha_{m}$ (Definition \autoref{def:reward_policy})                \\
		$\psi$            & $1e-4$      &    & Monotonically increasing H-parameter (Algorithm \autoref{algo:SEAC})                \\
		
		Optimizer         & Adam      &    & Refer to Adam \citep{kingma2014adam}                   \\
		environment steps & $1$       &    &              \\
		Replaybuffer size & $1e6$       &    &              \\
		Number of samples before training start & $5 \cdot max\_steps$       &    &              \\
		Number of critics & $2$       &    &              \\
		\bottomrule
	\end{tabular}
	\label{table:parameters}
\end{table}

\clearpage

\section{Statistic Results}
\label{sec:appendix_statistic}
	
	\begin{table}[h]
		\centering
		\caption{Paired Samples T-Test}
		\label{tab:pairedSamplesT-Test}
		\begin{tabular}{lrrrrr}
			\toprule
			Measure 1 &  & Measure 2 & W & z & p \\
			\cmidrule[0.4pt]{1-6}
			MOSEAC Energy Cost & - & SEAC Energy Cost & 2080.500 & -1.823 & 0.020 \\
			MOSEAC Time Cost & - & SEAC Time Cost & 18561.000 & -2.669 &  0.004 \\
			\bottomrule
		\end{tabular}
	\end{table}

\begin{table}[h]
	\centering
	\caption{Test of Normality (Shapiro-Wilk)}
	\label{tab:testOfNormality(Shapiro-Wilk)}
	\begin{tabular}{lrrrr}
		\toprule
		&  &  & W & p  \\
		\cmidrule[0.4pt]{1-5}
		MOSEAC Energy Cost & - & SEAC Energy Cost & 0.775 & < 0.001  \\
		MOSEAC Time Cost & - & SEAC Time Cost & 0.996 & 0.714  \\
		\bottomrule
	\end{tabular}
\end{table}

\begin{table}[h]
	\centering
	\caption{Descriptives}
	\label{tab:descriptives}
	\begin{tabular}{lrrrrr}
		\toprule
		& N & Mean & SD & SE & Coefficient of variation  \\
		\cmidrule[0.4pt]{1-6}
		MOSEAC Energy Cost & 300 & 3.120 & 0.424 & 0.024 & 0.136  \\
		SEAC Energy Cost & 300 & 3.193 & 0.451 & 0.026 & 0.141  \\
		MOSEAC Time Cost & 300 & 0.905 & 0.125 & 0.007 & 0.139  \\
		SEAC Time Cost & 300 & 0.935 & 0.127 & 0.007 & 0.136  \\
		\bottomrule
	\end{tabular}
\end{table}

\end{document}